# A Bayesian Optimization Algorithm for the Nurse Scheduling Problem

**Jingpeng Li**
School of Computer Science
University of Nottingham
NG8 1BB   UK
jpl@cs.nott.ac.uk

**Uwe Aickelin**
School of Computer Science
University of Nottingham
NG8 1BB   UK
uxa@cs.nott.ac.uk

**Abstract-** A Bayesian optimization algorithm for the nurse scheduling problem is presented, which involves choosing a suitable scheduling rule from a set for each nurse's assignment. Unlike our previous work that used GAs to implement implicit learning, the learning in the proposed algorithm is explicit, i.e. eventually, we will be able to identify and mix building blocks directly. The Bayesian optimization algorithm is applied to implement such explicit learning by building a Bayesian network of the joint distribution of solutions. The conditional probability of each variable in the network is computed according to an initial set of promising solutions. Subsequently, each new instance for each variable is generated by using the corresponding conditional probabilities, until all variables have been generated, i.e. in our case, a new rule string has been obtained. Another set of rule strings will be generated in this way, some of which will replace previous strings based on fitness selection. If stopping conditions are not met, the conditional probabilities for all nodes in the Bayesian network are updated again using the current set of promising rule strings. Computational results from 52 real data instances demonstrate the success of this approach. It is also suggested that the learning mechanism in the proposed approach might be suitable for other scheduling problems.

## 1 Introduction

Scheduling problems are generally NP-hard combinatorial problems, and a lot of research has been done to solve these problems heuristically (Aickelin and Dowsland, 2002 and 2003; Li and Kwan, 2001a and 2003). However, most previous approaches are problem-specific and research into the development of a general scheduling algorithm is still in its infancy.

Genetic Algorithms (GAs) (Holland 1975; Goldberg 1989), mimicking the natural evolutionary process of the survival of the fittest, have attracted much attention in solving difficult scheduling problems in recent years. Some obstacles exist when using GAs: there is no canonical mechanism to deal with constraints, which are commonly met in most real-world scheduling problems, and small changes to a solution are difficult. To overcome both difficulties, indirect approaches have been presented (Aickelin and Dowsland, 2003; Li and Kwan, 2001b and 2003) for nurse and driver scheduling. In these indirect GAs, the solution space is mapped and then a separate decoding routine builds solutions to the original problem.

In our previous indirect GAs, learning was implicit ('black-box') and restricted to the efficient adjustment of weights for a set of rules that are used to construct schedules. The major limitation of those approaches is that they learn in a non-human way. Like most existing construction algorithms, once the best weight combination is found, the rules used in the construction process are fixed at each iteration. However, normally a long sequence of moves is needed to construct a schedule and using fixed rules at each move is thus unreasonable and not coherent with the human learning processes.

When a human scheduler works, he normally builds a schedule systematically following a set of rules. After much practice, the scheduler gradually masters the knowledge of which solution parts go well with others. He can identify good parts and is aware of the solution quality even if the scheduling process is not completed yet, thus having the ability to finish a schedule by using flexible, rather than fixed, rules. In this paper, we design a more human-like scheduling algorithm, by using a Bayesian optimization algorithm to implement explicit learning from past solutions. A nurse scheduling problem with 52 real data instances gathered from a UK hospital is used as the test problem.

Nurse scheduling has been widely studied in recent years, and an extensive summary of the approaches can be found in Hung (1995) and Sitompul and Randhawa (1990). This problem is highly constrained, making it extremely difficult for most local search algorithms to find feasible solutions, let alone optimal ones. In our nurse scheduling problem, the number of the nurses is fixed (up to 30), and the target is to create a weekly schedule by assigning each nurse one out of up to 411 shift patterns in the most efficient way. The proposed Bayesian approach achieves this by choosing a suitable rule, from a rule set containing a number of available rules, for each nurse. A potential solution is therefore represented as a rule string, or a sequence of rules corresponding to nurses from the first one to the last.

As a model of the selected strings, a Bayesian network (Pearl 1998) is used in the proposed Bayesian optimization algorithm to solve the nurse scheduling problem. A Bayesian network is a directed acyclic graph with each node corresponding to one variable, and each variable corresponding to the individual rule by which a schedule will be constructed step by step. The causal relationship between two variables is represented by a directed edge between the two corresponding nodes.

The Bayesian optimization algorithm is applied to learn to identify good partial solutions and to complete them by building a Bayesian network of the joint distribution of solutions (Pelikan et al, 1999; Pelikan and Goldberg, 2000). The conditional probabilities are computed according to an initial set of promising solutions. Subsequently, each new instance for each node is generated by using the corresponding conditional probabilities, until values for all nodes have been generated, i.e. a new rule string has been generated.

Another set of rule strings will be generated in the same way, some of which will replace previous strings based on roulette-wheel fitness selection. If stopping conditions are not met, the conditional probabilities for all nodes in the Bayesian network are updated again using the current set of rule strings. The algorithm thereby tries to explicitly identify and mix promising building blocks.

It should be noted that for most scheduling problems, the structure of the network model is known and all variables are fully observed. In this case, the goal of learning is to find the rule values that maximize the likelihood of the training data. Thus, learning can amount to 'counting' in the case of multinomial distributions.

The rest of this paper is organized as follows. Section 2 gives an overview on the nurse scheduling problem, and the following section 3 introduces the general concepts about graphical models and Bayesian networks. Section 4 discuses the proposed Bayesian optimization algorithm, describing the construction of a Bayesian network, learning based on the Bayesian network, and the four building rules in detail. Computational results using 52 data instances gathered from a UK hospital are presented in section 5. Concluding remarks are in section 6.

## 2 The Nurse Scheduling Problem

### 2.1 General Problem

Our nurse scheduling problem is to create weekly schedules for wards of nurses by assigning one of a number of possible shift patterns to each nurse. These schedules have to satisfy working contracts and meet the demand for a given number of nurses of different grades on each shift, while being seen to be fair by the staff concerned. The latter objective is achieved by meeting as many of the nurses' requests as possible and considering historical information to ensure that unsatisfied requests and unpopular shifts are evenly distributed.

The problem is complicated by the fact that higher qualified nurses can substitute less qualified nurses but not vice versa. Thus scheduling the different grades independently is not possible. Furthermore, the problem has a special day-night structure as most of the nurses are contracted to work either days or nights in one week but not both. However due to working contracts, the number of days worked is not usually the same as the number of nights. Therefore, it becomes important to schedule the 'correct' nurses onto days and nights respectively. The latter two characteristics make this problem challenging for any local search algorithm, because finding and maintaining feasible solutions is extremely difficult.

The numbers of days or nights to be worked by each nurse defines the set of feasible weekly work patterns for that nurse. These will be referred to as shift patterns or shift pattern vectors in the following. For each nurse $i$ and each shift pattern $j$ all the information concerning the desirability of the pattern for this nurse is captured in a single numeric preference cost $p_{ij}$. These costs were determined in close consultation with the hospital and are a weighted sum of the following factors: basic shift-pattern cost, general day/night preferences, specific requests, continuity problems, number of successive working day, rotating nights/weekends and other working history information. Patterns that violate mandatory contractual requirements are marked as infeasible for a particular nurse and week by giving them a suitably high $p_{ij}$ value.

### 2.2 Integer Programming

The problem can be formulated as an integer linear program as follows.

Indices:
$i = 1...n$ nurse index;
$j = 1...m$ shift pattern index;
$k = 1...14$ day and night index (1...7 are days and 8...14 are nights);
$s = 1...p$ grade index.

Decision variables:
$$x_{ij} = \begin{cases} 1, & \text{nurse } i \text{ works shift pattern } j \\ 0, & \text{else} \end{cases}.$$

Parameters:
$m$ = Number of shift patterns;
$n$ = Number of nurses;
$p$ = Number of grades;
$$a_{jk} = \begin{cases} 1, & \text{shift pattern } j \text{ covers day/night } k \\ 0, & \text{else} \end{cases};$$
$$q_{is} = \begin{cases} 1, & \text{nurse } i \text{ is of grade } s \text{ or higher} \\ 0, & \text{else} \end{cases};$$
$p_{ij}$ = Preference cost of nurse $i$ working shift pattern $j$;
$R_{ks}$ = Demand of nurses with grade $s$ on day/night $k$;
$N_i$ = Working shifts per week of nurse $i$ if night shifts are worked;
$D_i$ = Working shifts per week of nurse $i$ if day shifts are worked;
$B_i$ = Working shifts per week of nurse $i$ if both day and night shifts are worked {for special nurses};
$F(i)$ = Set of feasible shift patterns for nurse $i$, where $F(i)$ is defined as

$$F(i) = \begin{cases} \sum_{k=1}^{7} a_{jk} = D_i, & \forall j \in \text{day shifts} \\ \sum_{k=8}^{14} a_{jk} = N_i, & \forall j \in \text{night shifts} \\ \sum_{k=1}^{14} a_{jk} = B_i, & \forall j \in \text{combined shifts} \end{cases}, \forall i. \quad (1)$$

Target function:
Minimize total preference cost of all nurses, denoted as

$$\sum_{i=1}^{n} \sum_{j \in F(i)}^{m} p_{ij} x_{ij} \to \min! \quad (2)$$

Subject to:
1. Every nurse works exactly one feasible shift pattern:
$$\sum_{j \in F(i)} x_{ij} = 1, \forall i; \quad (3)$$
2. The demand for nurses is fulfilled for every grade on every day and night:
$$\sum_{j \in F(i)} \sum_{i=1}^{n} q_{is} a_{jk} x_{ij} \geq R_{ks}, \forall k, s \quad (4)$$

Constraint set (3) ensures that every nurse works exactly one shift pattern from his/her feasible set, and constraint set (4) ensures that the demand for nurses is covered for every grade on every day and night. Note that the definition of $q_{is}$ is such that higher graded nurses can substitute those at lower grades if necessary.

Typical problem dimensions are 30 nurses of three grades and 400 shift patterns. Thus, the Integer Programming formulation has about 12000 binary variables and 100 constraints. This is a moderately sized problem. However, some problem cases remain unsolved after overnight computation using professional software.

## 3 Graphical Models and Bayesian Networks

In this section, we introduce concepts from graphical models in general and Bayesian networks in particular. Section 4 will then explain how we applied these concepts to our nurse scheduling problem.

Graphical models are graphs in which nodes represent random variables, and the lack of edges represents conditional independence assumptions (Edwards 2000). They have important applications in many multivariate probabilistic systems in fields such as statistics, systems engineering, information theory and pattern recognition. In particular, they are playing an increasingly important role in the design and analysis of machine learning algorithms.

As described by Jordon (1999), graphical models are a marriage between probability theory and graph theory. They provide a natural tool for dealing with uncertainty and complexity that occur throughout applied mathematics and engineering. In a graphical model, the fundamental notion of modularity is used to build a complex system by combining simpler parts. Probability theory provides the glue to combine the parts, ensuring that the whole system is consistent, and providing ways to interface models to data. The graph theory provides an intuitively appealing interface by which humans can model highly interacting sets of variables, and a data structure that leads itself naturally to the design of general-purpose algorithms.

There are two main kinds of graphical models: undirected and directed. Undirected graphical models are more popular with the physics and vision communities. Directed graphical model, also called Bayesian networks, are more popular with the artificial intelligence and machine learning communities. Bayesian networks are often used to model multinomial data with both discrete and continuous variables by encoding the relationship between the variables contained in the modelled data, which represents the structure of a problem.

Moreover, Bayesian networks can be used to generate new instances of the variables with similar properties as those of given data. Each node in the network corresponds to one variable, and each variable corresponds to one position in the strings representing the solutions. The relationship between two variables is represented by a directed edge between the two corresponding nodes.

Any complete probabilistic model of a domain must represent the joint distribution, the probability of every possible event as defined by the values of all the variables. The number of such events is exponential. To achieve compactness, Bayesian networks factor the joint distribution into local conditional distributions for each variable given its parents.

Mathematically, an acyclic Bayesian network encodes a full joint probability distribution by the product

$$P(x_1, ..., x_n) = \prod_{i=1}^{n} P(x_i \mid pa(X_i)) \quad , \quad (5)$$

where $x_i$ denotes some values of the variable $X_i$, $pa(X_i)$ denotes a set of values for parents of $X_i$ in the network (the set of nodes from which there exists an individual edge to $X_i$), and $P(x_i \mid pa(X_i))$ denotes the conditional probability of $X_i$ conditioned on variables $pa(X_i)$. This distribution can be used to generate new instances using the marginal and conditional probabilities.

## 4 A Bayesian Optimization Algorithm for Nurse Scheduling

This section discusses the proposed Bayesian optimization algorithm for the nurse scheduling problem, including the construction of a Bayesian network, learning based on the Bayesian network and the four building rules used.

### 4.1 The Construction of a Bayesian Network
In our nurse scheduling problem, the number of the nurse is fixed (up to 30), and the target is to create a weekly schedule by assigning each nurse one shift pattern in the most efficient way. The proposed approach achieves this by using one suitable rule, from a rule set that contains a number of available rules, for each nurse's assignment. Thus, a potential solution is represented as a rule string, or a sequence of rules corresponding to nurses from the first one to the last one individually.

We chose this approach, as the longer-term aim of our research is to model the explicit learning of a human scheduler. Human schedulers can provide high quality solutions, but the task is tedious and often requires a large amount of time. Typically, they construct schedules based on rules learnt during scheduling. Due to human limitations, these rules are typically simple. Hence, our rules will be relatively simple, too. Nevertheless, human generated schedules are of high quality due to the ability of the scheduler to switch between the rules, based on the state of the current solution. We envisage the Bayesian optimisation algorithm to perform this role.

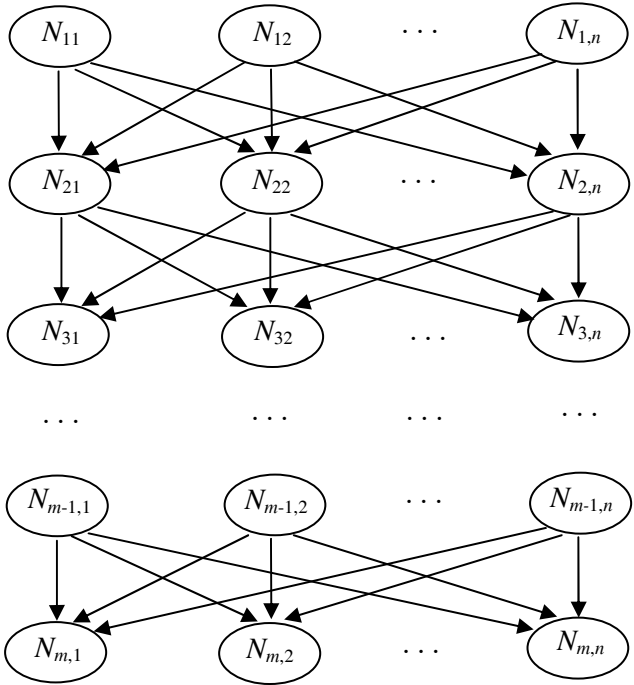

Figure 1: A Bayesian network for nurse scheduling

Figure 1 is the Bayesian network constructed for the nurse scheduling problem, which is a hierarchical and acyclic directed graph representing the solution structure of the problem.

The node $N_{ij} (i \in \{1,2,...,m\}; j \in \{1,2,...,n\})$ in the network denotes that nurse $i$ is assigned using rule $j$, where $m$ is the number of nurses to be scheduled and $n$ is the number of rules to be used in the building process. The directed edge from node $N_{ij}$ to node $N_{i+1,j'}$ denotes a causal relationship of "$N_{ij}$ causing $N_{i+1,j'}$". In our particular implementation, an edge denotes a construction unit (or rule sub-string) for nurse $i$ where the previous rule is $j$ and the current rule is $j'$. In this network, a possible solution (a complete rule string) is represented as a directed path from nurse 1 to nurse $m$ connecting $m$ nodes.

**4.2 Learning based on the Bayesian Network**

According to whether the structure (topology) of the model is known or unknown, and whether all variables are fully observed or some of them are hidden, there are four kinds of learning (Heckerman 1998). According to Heckerman, the learning process for the proposed approach belongs to the category of "known structure and full observation," and the learning goal is to find the variable values of all nodes $N_{ij}$ that maximize the likelihood of the training date containing $T$ independent cases.

In the proposed approach, learning amounts to counting and hence we use the symbol '#' meaning 'the number of' in the following equations. It calculates the conditional probabilities of each possible value for each node given all possible values of its parents. For example, for node $N_{i+1,j'}$ with a parent node $N_{ij}$, its conditional probability is

$$P(N_{i+1,j'} | N_{ij}) = \frac{P(N_{i+1,j'}, N_{ij})}{P(N_{ij})} \quad (6)$$

$$= \frac{\#(N_{i+1,j'} = true, N_{ij} = true)}{\#(N_{i+1,j'} = true, N_{ij} = true) + \#(N_{i+1,j'} = false, N_{ij} = true)}$$

Note that nodes $N_{1j}$ have no parents. In this circumstance, their probabilities are computed as

$$P(N_{1j}) = \frac{\#(N_{1j} = true)}{\#(N_{1j} = true) + \#(N_{1j} = false)} = \frac{\#(N_{1j} = true)}{T} \quad (7)$$

These probability values can be used to generate new rule strings, or new solutions. Since the first rule in a solution has no parents, it will be chosen from nodes $N_{1j}$ according to their probabilities. The next rule will be chosen from nodes $N_{ij}$ according to their probabilities conditioned on the previous nodes. This building process is repeated until the last node has been chosen from nodes $N_{mj}$, where $m$ is number of the nurses. A link from nurse 1 to nurse $m$ is thus created, representing a new possible solution. Since all the probability values are normalized, the roulette-wheel method is good strategy for rule selection.

For clarity, consider the following toy example of scheduling five nurses with two rules (1: random allocation, 2: allocate nurse to low-cost shifts). In the beginning of the search, the probabilities of choosing rule 1 or 2 for each nurse is equal, i.e. 50%. After a few iterations, due to the selection pressure and reinforcement learning, we experience two solution pathways: Because pure low-cost or random allocation produces low quality solutions, either rule 1 is used for the first 2-3 nurses and rule 2 on remainder or vice versa. In essence, BOA learns 'use rule 2 after 2-3x using rule 1' or vice versa.

**4.3 A Bayesian Optimization Algorithm**

Based on the estimation of conditional probabilities, this section introduces a Bayesian optimization algorithm for the nurse scheduling problem. It uses techniques from the field of modelling data by Bayesian networks to estimate the joint distribution of promising solutions. The nodes, or variables, in the Bayesian network correspond to the individual rules by which a schedule will be built step by step.

In the proposed Bayesian optimization algorithm, the first population of rule strings is generated at random. From the current population, a set of better rule strings is selected. Any selection method biased towards better fitness can be used, and in this paper, the traditional roulette-wheel Selection is applied. The conditional probabilities of each node in the Bayesian network are computed. New rule strings are generated by using these conditional probability values, and are added into the old population, replacing some of the old rule strings. In more detail, the steps of the Bayesian optimization algorithm for nurse scheduling are:

1. Set $t = 0$, and generate an initial population $P(0)$ at random;
2. Use roulette-wheel to select a set of promising rule strings $S(t)$ from $P(t)$;
3. Compute the conditional probabilities of each node according to this set of promising solutions ;
4. For the assignment of each nurse, the roulette-wheel method is used to select one rule according to the conditional probabilities of all available nodes, thus obtaining a new rule string. A set of new rule strings $O(t)$ will be generated in this way;
5. Create a new population $P(t+1)$ by replacing some rule strings from $P(t)$ with $O(t)$, and set $t = t+1$;
6. If the termination conditions are not met (we use 2000 generations), go to step 2.

**4.4 Four Building Rules**
Similar to the working pattern of a human scheduler, the proposed schedule-constructing process uses a set of rules to build a schedule step by step. As far as the domain knowledge of nurse scheduling is concerned, the following four rules are currently investigated.

4.4.1   Random Rule
The first rule, called 'Random' rule, is used to select a nurse's shift pattern at random. Its purpose is to introduce randomness into the search thus enlarging the search space, and most importantly to ensure that the proposed algorithm has the ability to escape from local optimum. This rule mirrors much of a scheduler's creativeness to come up with different solutions if required.

4.4.2   $k$-Cheapest Rule
The second rule is the '$k$-Cheapest' rule. Disregarding the feasibility of the schedule, it randomly selects a shift pattern from a $k$-length list containing patterns with $k$-cheapest cost $p_{ij}$, in an effort to reduce the total cost of a schedule as more as possible.

4.4.3   Cover Rule
Compared with the first two rules, the 'Cover' rule and last 'Contribution' rule are relatively more complicated. The third 'Cover' rule is designed to consider only the feasibility of the schedule. It schedules one nurse at a time in such a way as to cover those days and nights with the highest number of uncovered shifts.

The 'Cover' rule constructs solutions as follows. For each shift pattern in a nurse's feasible set, calculate the total number of uncovered shifts and would be covered if the nurse worked that shift pattern. For simplicity, this calculation does not take into account how many nurses are still required in a particular shift. For instance, assume that a shift pattern covers Monday to Friday nights. Further assume that the current requirements for the nights from Monday to Sunday are as follows: (-3, 0, +1, -2, -1, -2, 0), where a negative number means undercover and a positive over cover. The Monday to Friday shift pattern hence has a cover value of 3, as the most negative value it covers is -3. In this example, a Tuesday to Saturday pattern would have a value of 2.

In order to ensure that high-grade nurses are not 'wasted' covering unnecessarily for nurses of lower grades, for nurses of grade $s$, only the shifts requiring grade $s$ nurses are counted as long as there is a single uncovered shift for this grade. If all these are covered, shifts of the next lower grade are considered and once these are filled those of the next lower grade. Due to the nature of this approach, nurses' preference costs $p_{ij}$ are not taken into account by this rule. However, they will influence decisions indirectly via the fitness function. Hence, the 'Cover' rule can be summarised as finding those shift patterns with corresponding largest amount of undercover.

4.4.4   Contribution Rule
The fourth rule, called 'Contribution' rule, is biased towards solution quality but includes some aspects of feasibility by computing an overall score for each feasible pattern for the nurse currently being scheduled.

The 'Contribution' rule is designed to take into account the nurses' preferences. It therefore works with shift patterns rather than individual shifts. It also takes into account some of the covering constraints in which it gives preference to patterns that cover shifts that have not yet been allocated sufficient nurses to meet their total requirements. This is achieved by going through the entire set of feasible shift patterns for a nurse and assigning each one a score. The one with the highest (i.e. best) score is chosen. If there is more than one shift pattern with the best score, the first such shift pattern is chosen.

The score of a shift pattern is calculated as the weighted sum of the nurse's $p_{ij}$ value for that particular shift pattern and its contribution to the cover of all three grades. The latter is measured as a weighted sum of grade one, two and three uncovered shifts that would be covered if the nurse worked this shift pattern, i.e. the reduction in shortfall. Obviously, nurses can only contribute to uncovered demand of their own grade or below. More precisely and using the same notation as before, the score $p_{ij}$ of shift pattern $j$ for nurse $i$ is calculated with the following parameters:

- $d_{ks} = 1$ if there are still nurses needed on day $k$ of grade $s$ otherwise $d_{ks} = 0$;
- $a_{jk} = 1$ if shift pattern $j$ covers day $k$ otherwise $a_{jk} = 0$;
- $w_s$ is the weight of covering an uncovered shift of grade $s$;
- $w_p$ is the weight of the nurse's $p_{ij}$ value for the shift pattern.

Finally, $(100-p_{ij})$ must be used in the score, as higher $p_{ij}$ values are worse and the maximum for $p_{ij}$ is 100. Note that $(-w_p p_{ij})$ could also have been used, but would have led to some scores being negative. Thus, the scores are calculated as follows:

$$s_{ij} = w_p(100 - p_{ij}) + \sum_{s=1}^{3} w_s q_{is} (\sum_{k=1}^{14} a_{jk} d_{ks}) \quad (8)$$

The 'Contribution' rule can be summarised as follows:
- Cycle through all shift patterns of a nurse;
- Assign each one a score based on covering uncovered shifts and preference cost;
- Choose the shift pattern with the highest score.

### 4.5 Fitness Function

Independent of the rules used, the fitness of completed solutions has to be calculated. Unfortunately, feasibility cannot be guaranteed. This is a problem-specific issue and cannot be changed. Therefore, we still need a penalty function approach. Since the chosen encoding automatically satisfies constraint set (3) of the integer programming formulation, we can use the following formula, where $w_{demand}$ is the penalty weight, to calculate the fitness of solutions. Note that the penalty is proportional to the number of uncovered shifts.

$$\sum_{i=1}^{n}\sum_{j=1}^{m} p_{ij}x_{ij} + w_{demand}\sum_{k=1}^{14}\sum_{s=1}^{p} \max\left[R_{ks} - \sum_{i=1}^{n}\sum_{j=1}^{m} q_{is}a_{jk}x_{ij}; 0\right] \to \min! \quad (9)$$

## 5 Computational Results

In this section, we present the results of extensive computer experiments and compare them to results of the same data instances found previously by other algorithms. Table 1 lists the full and detailed computational results of 20 runs with different random seeds, where N/A indicates no feasible solution was found. Figures 2 summarises this information, Figure 3 shows a single typical run and finally Figure 4 gives an overall comparison between various algorithms.

### 5.1 Details of Algorithms

The results listed in Table 1 are always based on 20 runs with different random seeds and the last row contains the mean value of all columns:

- IP: Optimal or best-known solutions found with IP software (Dowsland and Thompson, 2000);
- GA: Best result out of 20 runs from a parallel genetic algorithm with multiple sub-populations and intelligent parameter adaptation (Aickelin and Dowsland, 2000);
- Rd: Bayesian optimization, but only the random rule is used, i.e. equivalent to random search;
- CP: Bayesian optimisation, where all four rules are used (see 4.4), but no conditional probability are computed, i.e. every rule has a 25% probability of being chosen all the time for all nurses;
- Op: Best result out of 20 runs of standard Bayesian optimization, i.e. four rules and conditional probabilities are used as described in section 4.1-4.4;
- Inf: Number of runs terminating with the best solution being infeasible;
- #: Number of runs terminating with the best solution being optimal or equal to the best known;
- <3: Number of runs terminating with the best solution being within three cost units of the optimum. The value of three units was chosen as it corresponds to the penalty cost of violating the least important level of requests in the original formulation. Thus, these solutions are still acceptable to the hospital.

For all data instances, the Bayesian optimisation algorithm used a set of fixed parameters as follows:
- Maximum number of generations = 2000;
- Penalty weight for each uncovered unit: $w_{demand}$ =200;
- For the '$k$-Cheapest' rule, $k$ = 5;
- Weight set for the 'Contribution' rule: w ={8,2,1,1};
- Population size = 140;
- Keep the best 40 solution in each generation;
- The executing time of the algorithm is approx. 10-20 seconds per run and data instance on a Pentium 4 PC.

N.B.: These fixed parameters are not necessarily the best for each instance. At this stage, there are based on our experience and intuition. We have kept them the same for consistency at this stage. When computing the mean a censored cost value of 255 has been used when an algorithm failed to find a feasible solution (N/A).

### 5.2 Analysis of Results

First, let us discuss the results in Table 1. Comparing the computational results on various test instances, one can see that using the random rule alone does not yield a single feasible solution. This underlines the difficulty of this problem. In addition, without learning the conditional probabilities, the results are much weaker, as the CP column shows. Thus, it is not simply enough to use the four rules to build solutions. Overall, the Bayesian results found rival those found by the complex multi-population GA. For some data instances, the results are much better. Particular impressive is the fact that in 100% of cases a feasible solution is found. Note that independent of the algorithm used, some data instances are harder to solve than others due to a shortage of nurses in some weeks.

| Set | IP | GA | Rd | CP | Op | Inf | # | <3 |
|---|---|---|---|---|---|---|---|---|
| 01 | 8 | 8 | N/A | 27 | 8 | 0 | 19 | 20 |
| 02 | 49 | 50 | N/A | 85 | 56 | 0 | 0 | 0 |
| 03 | 50 | 50 | N/A | 97 | 50 | 0 | 2 | 5 |
| 04 | 17 | 17 | N/A | 23 | 17 | 0 | 20 | 20 |
| 05 | 11 | 11 | N/A | 51 | 11 | 0 | 8 | 16 |
| 06 | 2 | 2 | N/A | 51 | 2 | 0 | 17 | 17 |
| 07 | 11 | 11 | N/A | 80 | 14 | 0 | 0 | 3 |
| 08 | 14 | 15 | N/A | 62 | 15 | 0 | 0 | 11 |
| 09 | 3 | 3 | N/A | 44 | 14 | 0 | 0 | 0 |
| 10 | 2 | 4 | N/A | 12 | 2 | 0 | 2 | 10 |
| 11 | 2 | 2 | N/A | 12 | 2 | 0 | 2 | 20 |
| 12 | 2 | 2 | N/A | 47 | 3 | 0 | 0 | 2 |
| 13 | 2 | 2 | N/A | 17 | 3 | 0 | 0 | 20 |
| 14 | 3 | 3 | N/A | 102 | 4 | 0 | 0 | 7 |
| 15 | 3 | 3 | N/A | 9 | 4 | 0 | 0 | 20 |
| 16 | 37 | 38 | N/A | 55 | 38 | 0 | 0 | 20 |
| 17 | 9 | 9 | N/A | 146 | 9 | 0 | 4 | 11 |
| 18 | 18 | 19 | N/A | 73 | 19 | 0 | 0 | 20 |
| 19 | 1 | 1 | N/A | 135 | 10 | 0 | 0 | 0 |
| 20 | 7 | 8 | N/A | 53 | 7 | 0 | 5 | 19 |
| 21 | 0 | 0 | N/A | 19 | 1 | 0 | 0 | 20 |
| 22 | 25 | 26 | N/A | 56 | 26 | 0 | 0 | 15 |
| 23 | 0 | 0 | N/A | 119 | 1 | 0 | 0 | 20 |
| 24 | 1 | 1 | N/A | 4 | 1 | 0 | 20 | 20 |
| 25 | 0 | 0 | N/A | 3 | 0 | 0 | 18 | 20 |
| 26 | 48 | 48 | N/A | 222 | 52 | 0 | 0 | 1 |

| | | | | | | | |
|---|---|---|---|---|---|---|---|
| 27 | 2 | 2 | N/A | 158 | 28 | 0 | 0 | 0 |
| 28 | 63 | 63 | N/A | 88 | 65 | 0 | 0 | 3 |
| 29 | 15 | 141 | N/A | 31 | 109 | 0 | 0 | 0 |
| 30 | 35 | 42 | N/A | 180 | 38 | 0 | 0 | 3 |
| 31 | 62 | 166 | N/A | 253 | 159 | 0 | 0 | 0 |
| 32 | 40 | 99 | N/A | 102 | 43 | 0 | 0 | 4 |
| 33 | 10 | 10 | N/A | 30 | 11 | 0 | 0 | 8 |
| 34 | 38 | 48 | N/A | 95 | 41 | 0 | 0 | 2 |
| 35 | 35 | 35 | N/A | 118 | 46 | 0 | 0 | 0 |
| 36 | 32 | 41 | N/A | 130 | 45 | 0 | 0 | 0 |
| 37 | 5 | 5 | N/A | 28 | 7 | 0 | 0 | 7 |
| 38 | 13 | 14 | N/A | 130 | 25 | 0 | 0 | 0 |
| 39 | 5 | 5 | N/A | 44 | 8 | 0 | 0 | 3 |
| 40 | 7 | 8 | N/A | 51 | 8 | 0 | 0 | 10 |
| 41 | 54 | 54 | N/A | 87 | 55 | 0 | 0 | 15 |
| 42 | 38 | 38 | N/A | 188 | 41 | 0 | 0 | 1 |
| 43 | 22 | 39 | N/A | 86 | 23 | 0 | 0 | 13 |
| 44 | 19 | 19 | N/A | 70 | 24 | 0 | 0 | 0 |
| 45 | 3 | 3 | N/A | 34 | 6 | 0 | 0 | 2 |
| 46 | 3 | 3 | N/A | 196 | 7 | 0 | 0 | 0 |
| 47 | 3 | 4 | N/A | 11 | 3 | 0 | 13 | 20 |
| 48 | 4 | 6 | N/A | 35 | 5 | 0 | 0 | 10 |
| 49 | 27 | 30 | N/A | 69 | 30 | 0 | 0 | 2 |
| 50 | 107 | 211 | N/A | 162 | 109 | 0 | 0 | 0 |
| 51 | 74 | N/A | N/A | 197 | 171 | 0 | 0 | 0 |
| 52 | 58 | N/A | N/A | 135 | 67 | 0 | 0 | 0 |
| *Av.* | **21** | **37** | **N/A** | **83** | **30** | **0** | **3** | **9** |

**Table 1: Comparison of results over 52 instances.**

Figures 2 and 3 show the results graphically. The bars above the *y*-axis represent solution quality. The black bars show the number of optimal, the grey near-optimal (within three units) solutions. The bars below the y-axis represent the number of times the algorithm failed to find a feasible solution. Hence, the shorter the bar is below the y-axis and the longer above, the better the algorithm's performance. Note that 'empty' bars mean that feasible, but not optimal solutions were found.

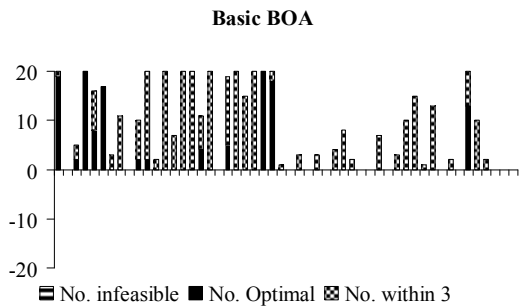

**Figure 2: The Bayesian optimisation algorithm.**

Figure 2 shows that for the Bayesian algorithm 38 out of 52 data sets are solved to or near to optimality. Additionally, feasible solutions are always found for all data sets and hence nothing is plotted below the x-axis.

For the GA in figure 3 the results are similar: 42 data sets are solved well, however many solutions are infeasible and for two instances not a single feasible solution had been identified. Both algorithms have difficulties solving the later data sets (nurse shortages), but BOA less so than the GA.

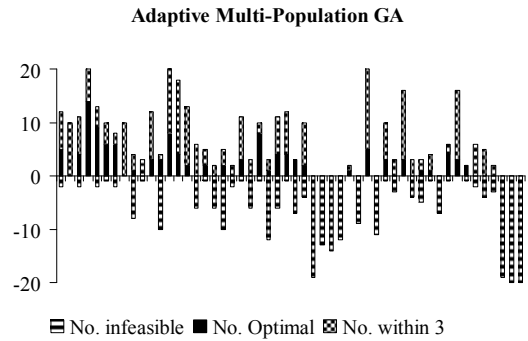

**Figure 3: The Genetic Algorithm.**

The behaviour of an individual run of the Bayesian algorithm is as expected. Figure 4 depicts the improvement of the schedule for the 04 data instance. At the generation of 57, the optimal solution cost 17 has been achieved. Although the actual values may differ among various instances, the characteristic shapes of the curves are similar for all seeds and data instances.

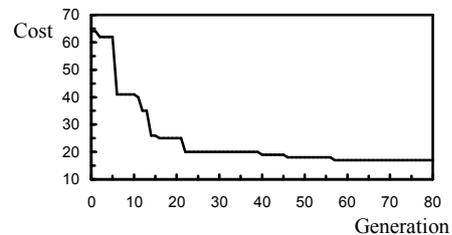

**Figure 4: Sample run of the Bayesian algorithm.**

Finally, Figure 5 compares performance of different GAs (Aickelin and Dowsland, 2000 and 2003) with the (Basic) Bayesian optimization algorithm presented here. The results are encouraging: with a fraction of the development time and simpler algorithm, the complex genetic algorithms are outperformed in terms of feasibility, best and average results.

Only the Hill-climbing GA, which includes an additional local search, has a better 'best case' performance. We believe that once this feature is added into the Bayesian optimization algorithm, we will see the best possible results. Our plan is to implement a post-processor that is similar to a human scheduler who 'improves' a finished schedule.

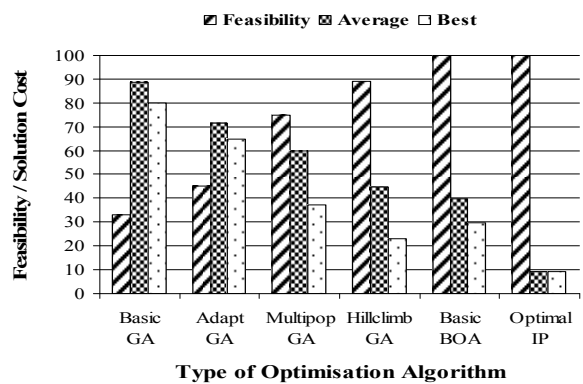

**Figure 5: Summary results of various algorithms.**

## 6 Conclusions

A new scheduling algorithm based on Bayesian networks is presented in this paper. The approach is novel because it is the first time that Bayesian networks have been applied to the field of personnel scheduling. An effective method is proposed to solve the problem about how to implement explicit learning from past solutions. Unlike most existing approaches, the new approach has the ability to build schedules by using flexible, rather than fixed rules. Experimental results from real-world nurse scheduling problems have demonstrated the strength of the proposed Bayesian optimization algorithm.

The proposed approach mimics human behaviour much more strongly than a standard GA based scheduling system. Although we have presented this work in terms of nurse scheduling, it is suggested that the main idea of the approach could be applied to many other scheduling problems where the schedules will be built systematically according to specific rules.

It is also hoped that this research will give some preliminary answers about how to include human-like learning into scheduling algorithms and may therefore be of interest to practitioners and researchers in areas of scheduling and evolutionary computation. In future, we will try to extract the 'explicit' part of the learning process further, e.g. by keeping partial solutions and learnt rules from one data instances to the next.

## Acknowledgements

The work was funded by the UK Government's major funding agency, Engineering and Physical Sciences Research Council (EPSRC), under grand GR/R92899/01.